# Defending from adversarial examples with a two-stream architecture

Hao Ge | Xiaoguang Tu | Mei Xie* | Zheng Ma

***Abstract***—In recent years, deep learning has shown impressive performance on many tasks. However, recent researches showed that deep learning systems are vulnerable to small, specially crafted perturbations that are imperceptible to humans. Images with such perturbations are the so called adversarial examples, which have proven to be an indisputable threat to the DNN based applications. The lack of better understanding of the DNNs has prevented the development of efficient defenses against adversarial examples. In this paper, we propose a two-stream architecture to protect CNN from attacking by adversarial examples. Our model draws on the idea of "two-stream" which commonly used in the security field, and successfully defends different kinds of attack methods by the differences of "high-resolution" and "low-resolution" networks in feature extraction. We provide a reasonable interpretation on why our two-stream architecture is difficult to defeat, and show experimentally that our method is hard to defeat with state-of-the-art attacks. We demonstrate that our two-stream architecture is robust to adversarial examples built by currently known attacking algorithms.

***Index Terms***— Deep learning, adversarial example, neural network

## 1. Introduction

With the development of CNN(Convolutional Neural Network), computers can deal with many tasks today, such as target classification[25-30], face recognition, license plate recognition, etc. In some tasks, computers performs even better than humans. Therefore, more and more works are done by computers rather than humans nowadays. However, just like hackers attack computer systems which threaten the security of computer, there are always unscrupulous people want to benefit from finding security holes in the security-sensitive fields, which impose secure threat to CNN-based systems.

So how about the security of the neural networks? It is a pity that, for almost every classification networks, lots of adversarial examples [2] can be generalized to mislead the classification result only by adding small perturbations on original images [3]. Such adversarial examples are potential threats to a wide range of applications (e.g. imagine that a "No passing" sign can be detected as a "No parking" sign by a self-driving car, just because of some small perturbations that humans are not aware of [4]). Therefore, finding a robust defensive method against adversarial attacks is really important nowadays.

The existing defense methods can be roughly divided into four categories: (1) Hiding the information of the target model to increase the difficulty of generating adversarial examples, e.g., defensive distillation [4],[10]; (2) Training the classifier with adversarial examples to improve its precision [3]; (3) Removing the adversarial perturbations by training a denoising autoencoder [5],[8]; (4) Training a classifier to distinguish between real images and adversarial examples [6],[7].

However, all these methods have disadvantages. For the first category, [20] showed that defensive distillation does not significantly increase the robustness of neural network; For categories 2 and 3, they need adversarial examples to train the defense, so these defenses are only effective to the process for generating those adversarial examples; For the last category, Carlini and Wagner [9] showed that these adversarial detecting methods can't defend their C&W Attack with slight changes on loss function.

Even for some powerful defense methods such as MagNet [7] and HGR [5], Carlini found them ineffective several days after they are published [22]. Viewing these challenges, we change our mind to build a defense system in a smaller scope to avoid being easily cracked.

Contributions: (1) We propose an efficient, effective defense method against adversarial examples. Our method is independent to the generation process of adversarial samples, as it requires only real images for training. (2) We explain the working mechanism of our "two-stream" method, which also explains why our method is difficult to attack.

## 2. Motivation

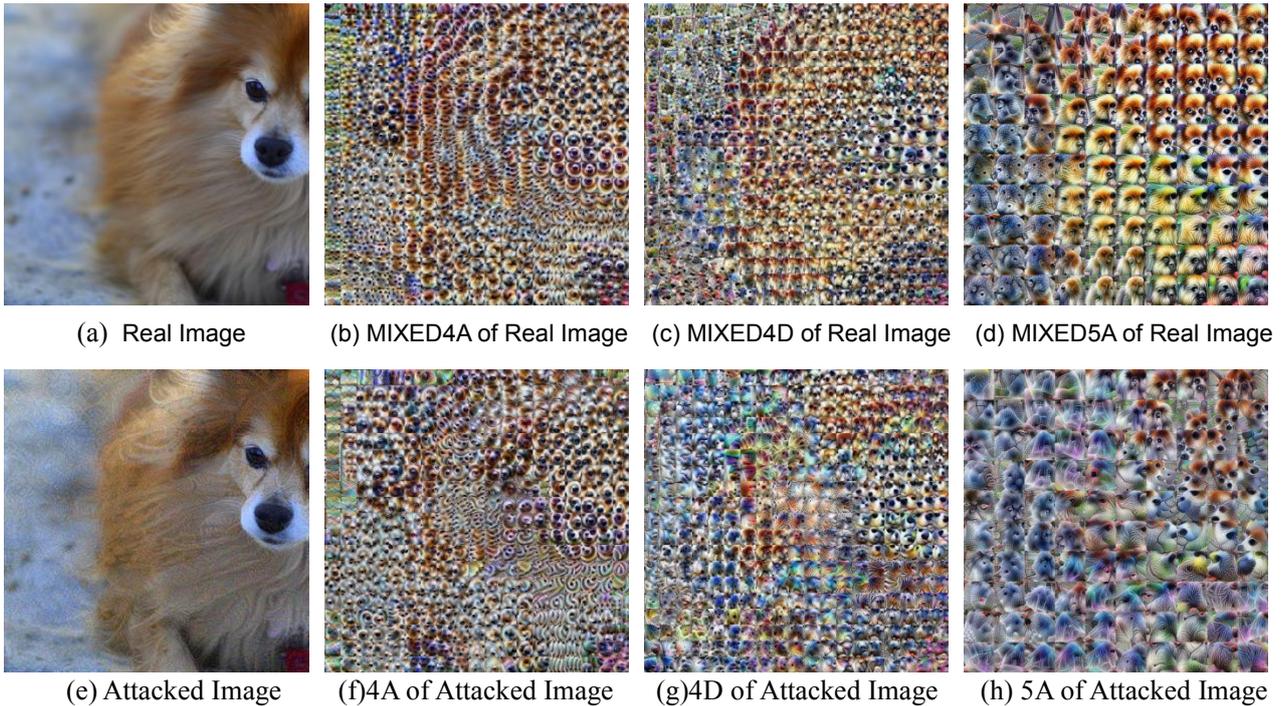

(a) Real Image  (b) MIXED4A of Real Image  (c) MIXED4D of Real Image  (d) MIXED5A of Real Image

(e) Attacked Image  (f) 4A of Attacked Image  (g) 4D of Attacked Image  (h) 5A of Attacked Image

Fig.1: Features visualization of the "Incv-3" network by Lucid [15].

As can be seen in Fig.1, we utilize Lucid to help us analyze what kind of errors the neural network made during the transmission of information when the input image is "adversarial example", which causes the final misclassification. The images from left to right in Fig.1 are the input images and the visualization results of each neuron in the 4A, 4D and 5A layers in GoogleNet. By observing Fig.1, we can find that, in the process of classifying a real image, the neurons in the "high-resolution" network can accurately classify the categories of the local areas according to the the texture information among the receptive fields, and these correct features can be delivered layer by layer, which leads to the correct classification result. However, as can be seen from the pictures on the second row, the existence of adversarial perturbations makes it impossible for the low-level neurons to accurately extract local features, which in turn affects the final classification results. This is the way how adversarial examples affect the "high-resolution" neural networks.

Here we have done a set of comparative experiments to explain why the neural network makes such mistakes from a human perspective in Fig.2. Fig.2 is obtained by dividing the two images in Fig.1 into 10*10 small squares respectively and then disturbing the arrangement of the small squares. The size of each small square in Fig.2 is approximately equal to the size of the receptive field of 4A layer in GoogleNet. In another word, all the information that each neuron in the 4A layer can obtain is included in a small square in Fig.2. And the disordered order is to make it impossible for us humans to judge the category by the context information of each small square, and only to look at each small square independently, thus the angle of view of each neuron node in the 4A layer is simulated. By carefully observing Fig.2, we can see that for the picture on the left, we humans can classify most of the small squares to "dog" without the outline information. However, when faced with the picture on the right, we will find that we are unable to accurately classify these small squares into "dog". The reason is that the perturbations destroy the texture features, so we can't accurately classify the small squares in the right picture by using the texture features just like what we do in the left picture.

The size of the small squares in Fig.2 is divided according to the size of the receptive field in the 4A layer in GoogleNet. Therefore, our neural network encounters the same problem in facing the adversarial examples. The change of texture features destroys the feature expression of each neuron, and such errors finally lead to a wrong classification result after being uploaded through layers. As to the real image in disordered order, the image on the left in Fig.2 can still be classified into "dog" with more than 90% confidence in GoogleNet, which confirms that the classification logic of GoogleNet, a high-resolution network, is different from humans, and they rely more on local features for image classification.

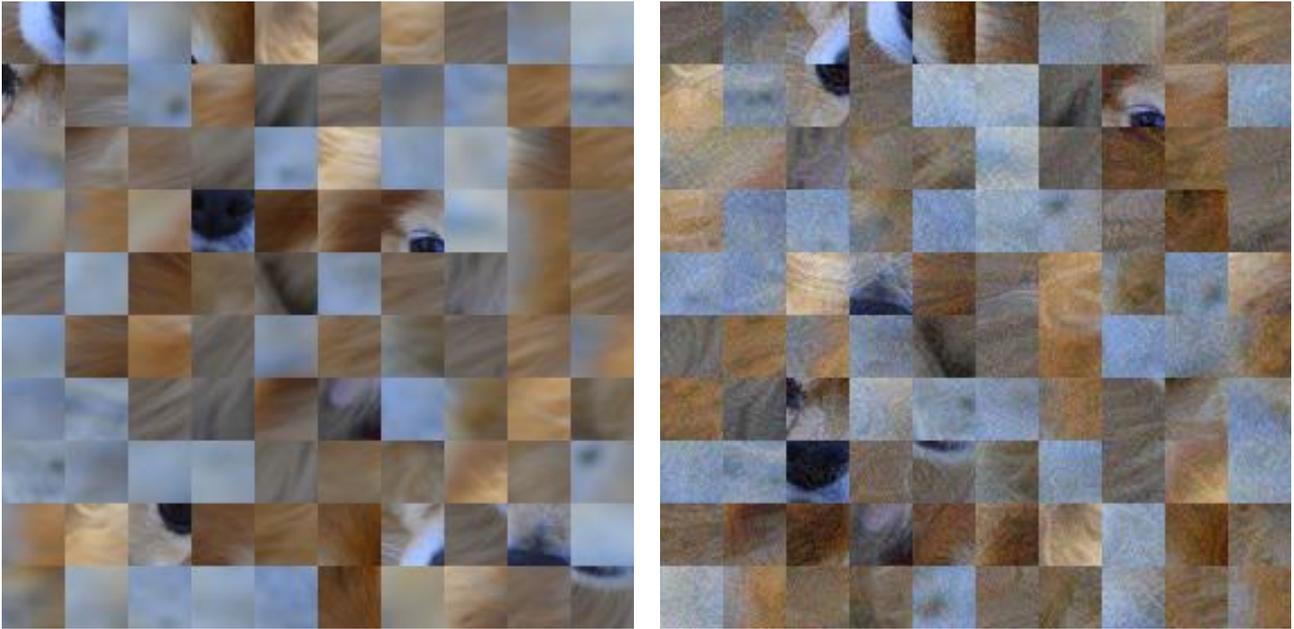

Fig.2: The Real image and the adversarial example in disordered order

Baker N [24] elaborated on the idea that neural networks do not rely on contour information in an article entitled "Deep convolutional networks do not classify based on global object shape". In this paper, the author combines the texture of the object A with the contour of the object B to test which feature the neural network is more dependent on. Reading this article will be of great help in reading our paper.

**2.1 What kind of problem has led to the failures of the defending methods?**

As mentioned above, many defending methods tends to fail against adversarial examples. It's difficult for these method to detect the adversarial examples [9], let alone correctly classification on them. What's the reason behind this situation? In this paper, we give the assumption that this reason is caused by the insufficient amount of data. From the perspective of information theory, all classification problems require a certain amount of information to support their classification results, and the information comes from the data involved in training. Adversarial perturbations increase the entropy of the pictures, so that the amount of information contained in the pictures are reduced, which lead to decrease of the information for correctly classification. And it is the decrease of information fails the defending methods. The adversarial perturbations degenerate the textures of the images. If we want to classify such images into right label, the defending methods should not depending on the texture features. To achieve this, a simple way is to enlarge the receptive field in CNNs, which is very similar to resizing the image to a smaller size. In our experiments, we did an experiment in ImageNet that resizing the training images to 32*32 to avoid interference from the adversarial perturbations, as a result, the testing accuracy is less than 10%, which proves that the amount of information in ImageNet is not sufficient to support 1000 categories of classification without the texture features. Similar to the situation in information theory that the coding algorithm with insufficient information will definitely fail, the classification tasks with insufficient information is like trying to make a dress with a handkerchief, which is destined to have a lot of loopholes.

Under the assumption that we can't obtain more data to offset the lack of information, we change our mind to build the defense system in a smaller scope to avoid of been cracked. Therefore, we propose a defense method that is extremely difficult to break under the following constraints: (1) The size of the input images should be 299*299, which is the size of the input images in GoogleNet. (2) The input images should involved in the 10 categories in CIFAR-10.

**2.2 Why choose "two-stream"?**

The idea of "two-stream" has been widely used in the security-sensitive field. For example, in the communication protocol, the "checksum" is used to transmit along with the "body part" to check for errors during transmission, the safe deposit box needs the keys of both the banker's and the customer's to open, important experiments need to be successfully replicated in different laboratories to be recognized, etc.

Moreover, during the research we found that the transferability of the adversarial examples are always pretty good when the target classifier is within googlenet, incv3, incv4, resnet and the networks derived from them. However, the fooling ratio will be much lower when the target classifier is CapsNet [23]. We believe that the reason of this phenomenon is that, the extraction of low-level features is more likely to be affected by the size of the receptive field of neurons in low layer. The low-layer neurons of the state-of-the-art classification algorithms have similar receptive fields, which leads to the similarity of the low-level features they extract, resulting in the transferability of the adversarial examples in these neural networks. However, the low-layer neurons in CapsNet have a much larger receptive field, which makes it more robust to the adversarial perturbations generated by the other networks.

In our "two-stream" architecture, the "low-resolution" network can be treat as a network with large receptive field for low layer neurons in dealing with high-resolution images, so the transferability of the adversarial examples between "high-resolution" and "low-resolution" network is bad, which is the reason why our method is effective.

## 3.    Our method

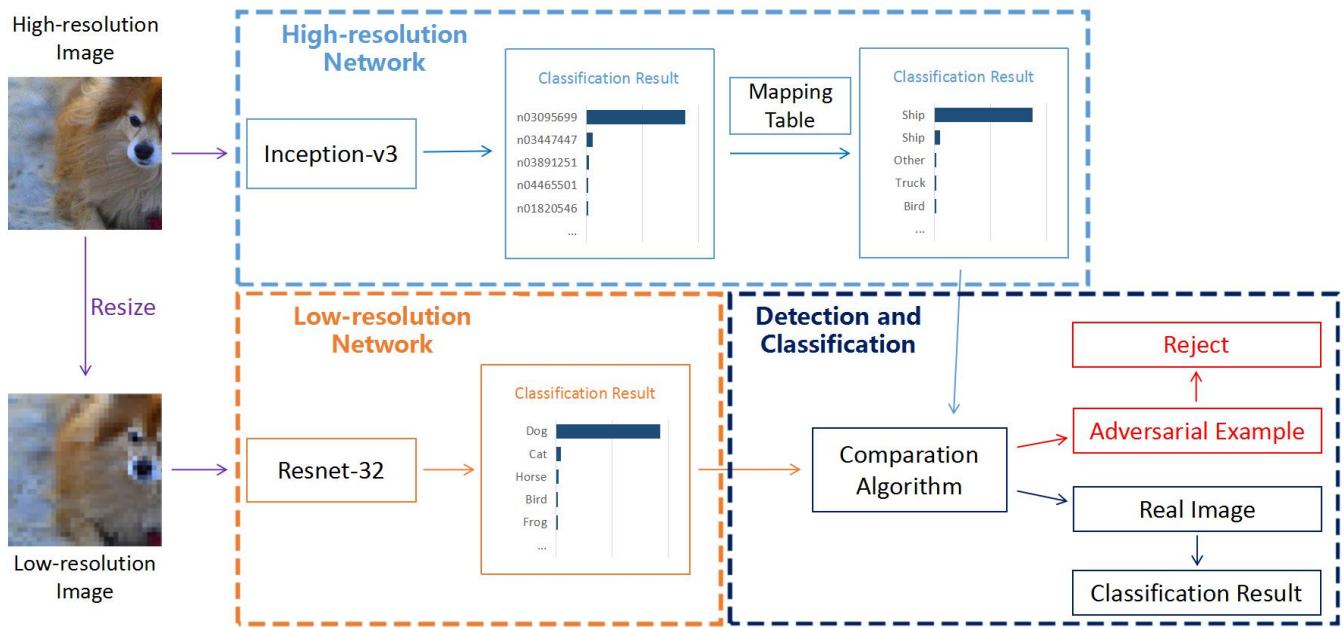

Fig.3: The framework of our two-stream network

Similar with SafetyNet [19] and MagNet [7], the workflow of our "two-stream" architecture consists of two steps: (1) a detector that rejects the adversarial examples and (2) a classifier that classifies the remaining images to the right label. As shown in Fig.3, the classification results of the "high-resolution" and "low-resolution" networks are fed to the comparation algorithm, and the comparation algorithm acts as both a detector and a classifier. The specific comparation algorithm is shown in Algorithm 1. The Mapping Table is a Table that maps the labels in Imagenet to the labels of Cifar-10, e.g., n02123045, n02124075,... → "Cat"; n02110063, n02110806,... → "Dog", etc.

It is worth noting that this is just a generic backbone, and the network used in this framework can be replace by other backbones (e.g. the Incv3 in "high-resolution" network can be replaced by VGG16, ResNet-152 or the other networks trained in ImageNet, and the ResNet-32 in "low-resolution" network can also be replaced by NiN, AllConv or the other networks trained in Cifar-10), which greatly increases the flexibility of this framework, making it difficult for attackers to implement white box attacks.

Algorithm 1 is the comparation algorithm, where p_1 and p_2 are hyperparameters as thresholds, which are set to 10% and 20% in our experiments, respectively. Y indicates the Labels, and P indicates the probability of these Labels. $Y_{high}$ and $P_{high}$ indicate the label and its corresponding probability of the TOP-5 classification results of the "High-resolution" network, $Y_{low}$ and $P_{low}$ indicate same things of the "Low-resolution" network.

**Algorithm 1** Classification

**Input:** $X_n$

**Output:** Classification Result $Y_n$

The TOP-5 classification results of the "high-resolution" network for image $X_n$ is set to be: $[(Y_{high}^{n(1)}, P_{high}^{n(1)}), \ldots (Y_{high}^{n(5)}, P_{high}^{n(5)})]$

The TOP-2 classification results of the "low-resolution" network for image $X_n$ is set to be: $[(Y_{low}^{n(1)}, P_{low}^{n(1)}), (Y_{low}^{n(2)}, P_{low}^{n(2)})]$

**for** n = 1 to N **do**
    initialize $i_0 \leftarrow 0, j_0 \leftarrow 0$
    **for all** $Y_{high}^{n(i)} = Y_{low}^{n(j)}$ **do**
        **for all** $Y_{high}^{n(i)} \geqslant p_1$ and $Y_{low}^{n(j)} \geqslant p_2$ **do**
            $i_0, j_0 \leftarrow argmax(Y_{high}^{n(i)} * Y_{low}^{n(j)})$
        **end for**
    **end for**
    **if** $i_0 = j_0 = 0$ **then**
        $Y_n \leftarrow$ Adversarial Example
    **else**
        $Y_n \leftarrow Y_{high}^{n(i_0)}$
    **end if**
**end for**

In order to verify the practicality of our proposed method, we built a network of 10000 user nodes and 1 server node to simulate a real network environment. The user node consists of 9000 normal user nodes and 1000 adversarial user nodes. Each normal user node periodically sends a real picture to the server to request the classification result, and the adversarial user node periodically sends an adversarial example. What the server node needs to do is to find and add these adversarial user nodes to the blacklist to prevent them from accessing the server, and return the correct classification results to the normal user nodes at the same time. In order to achieve this goal, we adopt the following algorithm 2 on the server node to distinguish whether a user node should be blacklisted. The sources of the images are recorded in $Img_n^{IP}$, The confidence coefficient for each IP is set to CC[IP], CC[IP] $\in$ [0, 15], The blacklist is set to be Bl[], The detection result of our "two-stream" network is recorded in $Img_n^{RoF}$, "1" means it's a real image, and "0" means not.

**Algorithm 2** Real or Fake

**Input:** images received by the server node: $Img_n$, the results of our "two-stream" classifier for $Img_n$: $Y_n$

**Output:** The result sent back to the user node for $Img_n$: $Cls_n$

Initialize Conf[] = Bl[] = []

**For** n = 1 to N **do**
    **if** $Img_n^{RoF} = 0$ **then**
        CC[IP]++
    **else**
        CC[IP]- -
    **end if**
    **if** $Img_n^{IP} \in$ Bl[] **then**
        **if** CC[IP] = 0 **then**
            Remove $Img_n^{IP}$ from Bl[], $Cls_n \leftarrow Y_n$
        **end if**
    **else**
        **if** CC[IP] $\geq$ 3 **then**
            add $Img_n^{IP}$ to Bl[], CC[IP] $\leftarrow Y_n$

```
            else
                Cls_n ← Y_n
            end if
        end if
    end for
```

## 4. Attacking Methods

In this paper, we divide the attack methods into two categories, namely type I attack and type II attack. Type I attack aims at fooling the high-resolution network and type II attack aims at the low-resolution network. We evaluated our defense against four popular attacks. Universal Adversarial Perturbations is type I attack, One Pixel Attack and Carlini Attack are type II attacks, FGSM can be both type I and type II attack. We now explain these attacks one by one.

**Fast Gradient Sign Method (FGSM)**: Goodfellow et al. [3] introduced this adversarial attack algorithm. They developed a method to generate an adversarial example by solving the following problem: $x' = x + \varepsilon \cdot sign(\nabla_x Loss(x, l_x))$. This attack is simple, but effective. Kurakin et al. [17] described an iteration version of the FGSM. For each iteration, the attack applies FGSM with a small step size α and clips the updated result after each iteration so that the updated image stays in the ε neighborhood of the original image. However, this adversarial attack can hardly fool a black-box model. To address this issue, Dong Y [18] proposed momentum iterative fast gradient sign method (MI-FGSM) to boost adversarial attacks.

**Universal Adversarial Perturbations**: Followed their previous work on DeepFool [16], Moosavi-Dezfooli [1] proposed this universal adversarial attack. Unlike other methods that compute perturbations to fool a network on a single image, this method is able to fool a network on all images. Moreover, the universal perturbations were shown to be generalized well across different neural networks.

**One Pixel Attack**: Su J [14] introduced this adversarial attack algorithm. They generate adversarial examples by only modifying one pixel. They claimed successful fooling of three common deep neural network on about 70% of the tested images. It is worth noting that, this attack method generates adversarial examples without any information about the parameter values or the gradients of the network. We utilize the "one-pixel" and "three-pixel" versions to test our method in our experiment.

**Carlini Attack**: Carlini [21] introduced an attack method for Cifar-10 and MNIST. It is the most powerful type II attack we found.

## 5. Evaluation

We evaluate the properties of our "two-stream" architecture on three datasets: car196 [12], fgvc-aircraft [13] and ImageNet [11]. Car196 [12] and fgvc-aircraft [13] are fine-grained datasets, which cantains 16185 images of 196 classes of cars and 10200 images of 102 kind of aircrafts, respectively. In this paper, these two databases are used to test the defensive performance of our architecture for the "automobile", "truck" and "airplane" categories. And the Imagenet [11] used in this paper is composed of the Cifar-10 related categories selected from the original Imagenet database, e.g. n01582220, n01601694$→bird, n01644373, n01644900→frog. There are totally 217 out of 1000 categories in Imagenet that can be classified into the 10 categories in Cifar-10, and the other 783 are labeled by "other".

The classification results of the "High-resolution" and "Low-resolution" networks are directly used to determine whether an image is an adversarial example or not, so it will be a disaster for our framework if there is an attack method that can take effects on both networks. To this end, we did an experiment to test the performance of the state-of-the-art attack algorithms on both of the networks. The experimental results are shown on Table 1, and "Non Attack Data" is a control group here.

In Table 1, H-Net means the Top-5 accuracy of the "High-resolution" network. L-Net means the Top-1 accuracy of the "Low-resolution" network. For the type I attack on Cifar-10, we resized the images from 32*32 to 299*299 to make them can be attacked by the type I attacks like the high-resolution datasets. And the process shown in Fig.4 is used to achieve the type II attacks on high-resolution datasets.

Table 1: Classification accuracy of "High-resolution" and "Low-resolution" networks on adversarial examples generated by different attack methods.

|  | Method | Cifar-10 | | Car196 [12] | | fgvc-aircraft [13] | | ImageNet [11] | |
|---|---|---|---|---|---|---|---|---|---|
|  |  | H-Net | L-Net | H-Net | L-Net | H-Net | L-Net | H-Net | L-Net |
|  | Non Attack Data | N/A | 91.4% | 95.6% | 91.9% | 98.9% | 94.3% | 96.7% | 87.1% |
| Type I Attack | FGSM [3] | N/A | 88.8% | 29.1% | 91.3% | 40.9% | 94.1% | 27.7% | 86.6% |
|  | I-FGSM [17] | N/A | 88.9% | 0.0% | 91.8% | 0.0% | 93.9% | 0.0% | 87.0% |
|  | MI-FGSM [18] | N/A | 88.8% | 0.0% | 92.2% | 0.0% | 94.0% | 0.0% | 86.5% |
|  | DeepFool [16] | N/A | 88.7% | 66.8% | 91.7% | 79.3% | 93.9% | 67.6% | 86.7% |
|  | Universal [1] | N/A | 88.6% | 34.8% | 91.8% | 49.3% | 94.2% | 26.0% | 86.8% |
| Type II Attack | FGSM | N/A | 40.2% | 95.3% | 54.2% | 98.7% | 72.8% | 96.6% | 37.5% |
|  | I-FGSM | N/A | 11.4% | 95.2% | 13.4% | 98.7% | 25.1% | 96.5% | 11.0% |
|  | One-pixel [14] | N/A | 36.5% | 95.5% | 44.5% | 99.1% | 86.2% | 96.4% | 32.3% |
|  | Three-pixel [14] | N/A | 13.5% | 95.5% | 26.0% | 99.1% | 55.1% | 96.4% | 13.1% |
|  | Carlini [21] | N/A | 0.0% | 95.1% | 0.0% | 98.3% | 0.0% | 96.4% | 0.0% |

As can be seen from Table 1, Type I Attack can only affect the classification result of "H-Net", and Type II Attack can only affect the classification result of "L-Net", in another word, Neither type I attack nor type II attack can be effective on both networks, which means that their misclassification results are irrelevant, so it is feasible to determine whether the input image is adversarial example by comparing their classification results. In addition, it is worth mentioning that it is not the attacking methods that caused the accuracy to drop while attacking Cifar-10 with type I attacks. We did a comparative experiment that just resize the images in Cifar-10 to 299*299 and then resize back to 32*32, the classification result of these images is 88.9%, which means that it is the "resize" but not the attack methods that reduced the accuracy.

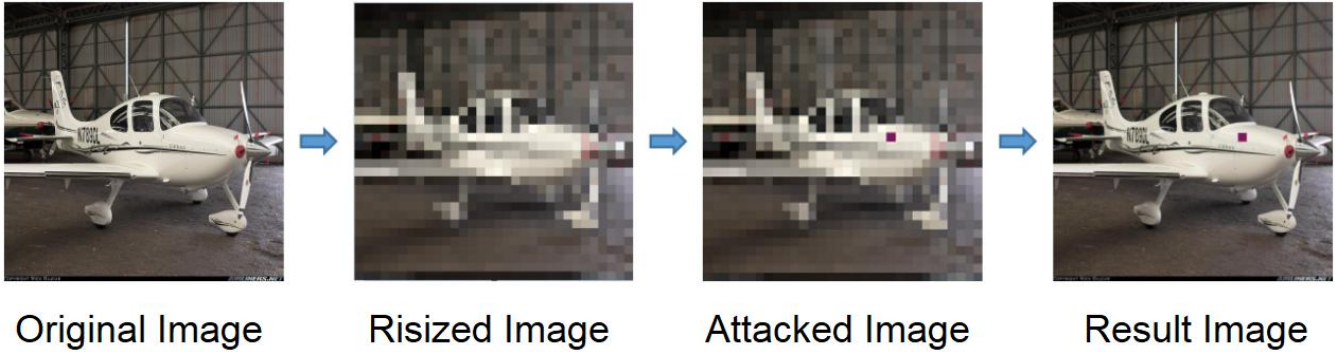

Fig.4: The flowchart for implementing the Type II Attack in high-resolution images

For the Type II Attack in high-resolution images, we take three steps to achieve the attack, the image is resized to 32*32 first, then this resized image is attacked by Type II Attack, and then the difference between the obtained adversarial example and the 32*32 original image is calculated, finally, the difference map is zoom in and overlay into the original image.

Table 2 shows the detection and classification results of our "two-stream" architecture. "Reject" means the rate at which the images are detected as adversarial examples and rejected by our "two-stream" architecture. "Right" means the rate at which the images are not rejected and classified into the right label, same for "Wrong". As can be seen from Table 2, almost all of the images can be either detected as "adversarial example" or classified into the right label. In another word, it is hard to produce an example that (a) is mislabeled and (b) is not detected as an adversarial example by "two-stream" architecture. And this is the standard Lu J [19] proposed to evaluate the quality of a defense method.

The experimental results of simulating a real-world network environment are shown in Fig.5, each polyline represents the proportion of a class of user nodes that are blacklisted. The horizontal axis represents the number of images sent by the user node and the vertical axis represents the proportion of blacklisted.

Table 2: Summary of the reaction of our "two-stream" architecture on various attacks.

| | Method | car196 [12] | | | fgvc-aircraft [13] | | | ImageNet [11] | | |
|---|---|---|---|---|---|---|---|---|---|---|
| | | Reject | Right | Wrong | Reject | Right | Wrong | Reject | Right | Wrong |
| | Non Attack Data | 11.0% | 88.9% | 0.1% | 6.4% | 93.6% | 0.0% | 13.3% | 86.6% | 0.1% |
| Type I Attack | FGSM [3] | 80.8% | 18.6% | 0.6% | 70.8% | 28.9% | 0.3% | 83.6% | 15.5% | 0.9% |
| | I-FGSM [17] | 99.1% | 0.0% | 0.9% | 99.3% | 0.0% | 0.7% | 98.5% | 0.0% | 1.5% |
| | MI-FGSM [18] | 99.1% | 0.0% | 0.9% | 92.2% | 0.0% | 0.8% | 98.5% | 0.0% | 1.5% |
| | DeepFool [16] | 41.1% | 58.6% | 0.3% | 31.7% | 68.2% | 0.1% | 46.4% | 53.2% | 0.4% |
| | Universal [1] | 79.9% | 19.6% | 0.5% | 66.4% | 33.3% | 0.3% | 83.7% | 15.3% | 1.0% |
| Type II Attack | FGSM | 42.9% | 56.8% | 0.3% | 25.2% | 74.6% | 0.2% | 61.4% | 38.2% | 0.4% |
| | I-FGSM | 84.3% | 15.2% | 0.5% | 73.5% | 26.1% | 0.4% | 88.1% | 11.3% | 0.6% |
| | One-pixel [14] | 49.9% | 49.7% | 0.3% | 10.7% | 89.2% | 0.1% | 65.5% | 34.1% | 0.4% |
| | Three-pixel [14] | 67.7% | 31.9% | 0.4% | 35.9% | 63.9% | 0.2% | 85.6% | 13.9% | 0.5% |
| | Carlini [21] | 99.1% | 0.3% | 0.6% | 99.2% | 0.4% | 0.4% | 98.9% | 0.4% | 0.7% |

It can be seen that the adversarial user nodes with strong perturbations(Incv3, Universal) are rapidly blacklisted, and the nodes with weak perturbations also have a high probability of being blacklisted, meanwhile, referring to Table 2, the classification results returned to these unshielded nodes are often the correct classification results. Therefore, not our defending algorithm is not strong enough, but the strength of these attack algorithms(DeepFool, Three-pixel) are not enough for us to shield them. As to the normal user nodes, the proportion almost equal to 0. Actually, during our totally 50 times experiments, only 17 normal user-nodes were added to the blacklist. Therefore, our defense algorithm performs well for both single images and simulated real-world network environments.

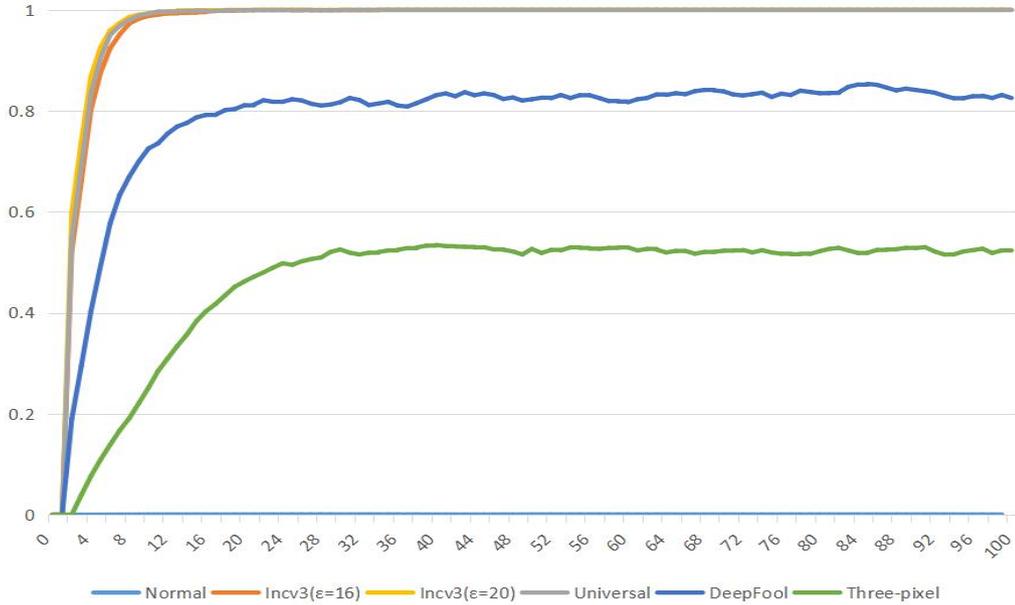

Fig.5: The proportion of user nodes being blacklisted by the server node

## 6.   Conclusion

In this paper, we propose a new "two-stream" architecture to defend against adversarial examples. By comparing the classification results of the "high-resolution" and "low-resolution" networks, our "two-stream" framework is able to detect adversarial examples without requiring either adversarial examples or the knowledge of the generation process. It is two kinds of networks but not two specific networks that compared in our "two-stream" framework, which makes it: (1) can be further enhanced by new datasets and new backbones in the future, (2) difficult for an attacker to implement white-box attack. Experiments show that, it is hard to produce an example that (a) is mislabeled and (b) is not detected as an adversarial example by "two-stream" architecture. Moreover, we sketched one possible reason for why "two-stream" works by analyzing the impact of adversarial perturbations on neural networks.